\documentclass[11pt]{article}

\usepackage{colacl}

\makeatletter
\newcounter{treecount}
\newcounter{branchcount}
\newsavebox{\parentbox}
\newsavebox{\treebox}
\newsavebox{\treeboxone}
\newsavebox{\treeboxtwo}
\newsavebox{\treeboxthree}
\newsavebox{\treeboxfour}
\newsavebox{\treeboxfive}
\newsavebox{\treeboxsix}
\newsavebox{\treeboxseven}
\newsavebox{\treeboxeight}
\newsavebox{\treeboxnine}
\newsavebox{\treeboxten}
\newsavebox{\treeboxeleven}
\newsavebox{\treeboxtwelve}
\newsavebox{\treeboxthirteen}
\newsavebox{\treeboxfourteen}
\newsavebox{\treeboxfifteen}
\newsavebox{\treeboxsixteen}
\newsavebox{\treeboxseventeen}
\newsavebox{\treeboxeighteen}
\newsavebox{\treeboxnineteen}
\newsavebox{\treeboxtwenty}
\newlength{\treeoffsetone}
\newlength{\treeoffsettwo}
\newlength{\treeoffsetthree}
\newlength{\treeoffsetfour}
\newlength{\treeoffsetfive}
\newlength{\treeoffsetsix}
\newlength{\treeoffsetseven}
\newlength{\treeoffseteight}
\newlength{\treeoffsetnine}
\newlength{\treeoffsetten}
\newlength{\treeoffseteleven}
\newlength{\treeoffsettwelve}
\newlength{\treeoffsetthirteen}
\newlength{\treeoffsetfourteen}
\newlength{\treeoffsetfifteen}
\newlength{\treeoffsetsixteen}
\newlength{\treeoffsetseventeen}
\newlength{\treeoffseteighteen}
\newlength{\treeoffsetnineteen}
\newlength{\treeoffsettwenty}

\newlength{\treeshiftone}
\newlength{\treeshifttwo}
\newlength{\treeshiftthree}
\newlength{\treeshiftfour}
\newlength{\treeshiftfive}
\newlength{\treeshiftsix}
\newlength{\treeshiftseven}
\newlength{\treeshifteight}
\newlength{\treeshiftnine}
\newlength{\treeshiftten}
\newlength{\treeshifteleven}
\newlength{\treeshifttwelve}
\newlength{\treeshiftthirteen}
\newlength{\treeshiftfourteen}
\newlength{\treeshiftfifteen}
\newlength{\treeshiftsixteen}
\newlength{\treeshiftseventeen}
\newlength{\treeshifteighteen}
\newlength{\treeshiftnineteen}
\newlength{\treeshifttwenty}
\newlength{\treewidthone}
\newlength{\treewidthtwo}
\newlength{\treewidththree}
\newlength{\treewidthfour}
\newlength{\treewidthfive}
\newlength{\treewidthsix}
\newlength{\treewidthseven}
\newlength{\treewidtheight}
\newlength{\treewidthnine}
\newlength{\treewidthten}
\newlength{\treewidtheleven}
\newlength{\treewidthtwelve}
\newlength{\treewidththirteen}
\newlength{\treewidthfourteen}
\newlength{\treewidthfifteen}
\newlength{\treewidthsixteen}
\newlength{\treewidthseventeen}
\newlength{\treewidtheighteen}
\newlength{\treewidthnineteen}
\newlength{\treewidthtwenty}
\newlength{\daughteroffsetone}
\newlength{\daughteroffsettwo}
\newlength{\daughteroffsetthree}
\newlength{\daughteroffsetfour}
\newlength{\branchwidthone}
\newlength{\branchwidthtwo}
\newlength{\branchwidththree}
\newlength{\branchwidthfour}
\newlength{\parentoffset}
\newlength{\treeoffset}
\newlength{\daughteroffset}
\newlength{\branchwidth}
\newlength{\parentwidth}
\newlength{\treewidth}
\newcommand{\ontop}[1]{\begin{tabular}{c}#1\end{tabular}}
\newcommand{\poptree}{%
\ifnum\value{treecount}=0\typeout{QobiTeX warning---Tree stack underflow}\fi%
\addtocounter{treecount}{-1}%
\setlength{\treeoffsettwo}{\treeoffsetthree}%
\setlength{\treeoffsetthree}{\treeoffsetfour}%
\setlength{\treeoffsetfour}{\treeoffsetfive}%
\setlength{\treeoffsetfive}{\treeoffsetsix}%
\setlength{\treeoffsetsix}{\treeoffsetseven}%
\setlength{\treeoffsetseven}{\treeoffseteight}%
\setlength{\treeoffseteight}{\treeoffsetnine}%
\setlength{\treeoffsetnine}{\treeoffsetten}%
\setlength{\treeoffsetten}{\treeoffseteleven}%
\setlength{\treeoffseteleven}{\treeoffsettwelve}%
\setlength{\treeoffsettwelve}{\treeoffsetthirteen}%
\setlength{\treeoffsetthirteen}{\treeoffsetfourteen}%
\setlength{\treeoffsetfourteen}{\treeoffsetfifteen}%
\setlength{\treeoffsetfifteen}{\treeoffsetsixteen}%
\setlength{\treeoffsetsixteen}{\treeoffsetseventeen}%
\setlength{\treeoffsetseventeen}{\treeoffseteighteen}%
\setlength{\treeoffseteighteen}{\treeoffsetnineteen}%
\setlength{\treeoffsetnineteen}{\treeoffsettwenty}%
\setlength{\treeshifttwo}{\treeshiftthree}%
\setlength{\treeshiftthree}{\treeshiftfour}%
\setlength{\treeshiftfour}{\treeshiftfive}%
\setlength{\treeshiftfive}{\treeshiftsix}%
\setlength{\treeshiftsix}{\treeshiftseven}%
\setlength{\treeshiftseven}{\treeshifteight}%
\setlength{\treeshifteight}{\treeshiftnine}%
\setlength{\treeshiftnine}{\treeshiftten}%
\setlength{\treeshiftten}{\treeshifteleven}%
\setlength{\treeshifteleven}{\treeshifttwelve}%
\setlength{\treeshifttwelve}{\treeshiftthirteen}%
\setlength{\treeshiftthirteen}{\treeshiftfourteen}%
\setlength{\treeshiftfourteen}{\treeshiftfifteen}%
\setlength{\treeshiftfifteen}{\treeshiftsixteen}%
\setlength{\treeshiftsixteen}{\treeshiftseventeen}%
\setlength{\treeshiftseventeen}{\treeshifteighteen}%
\setlength{\treeshifteighteen}{\treeshiftnineteen}%
\setlength{\treeshiftnineteen}{\treeshifttwenty}%
\setlength{\treewidthtwo}{\treewidththree}%
\setlength{\treewidththree}{\treewidthfour}%
\setlength{\treewidthfour}{\treewidthfive}%
\setlength{\treewidthfive}{\treewidthsix}%
\setlength{\treewidthsix}{\treewidthseven}%
\setlength{\treewidthseven}{\treewidtheight}%
\setlength{\treewidtheight}{\treewidthnine}%
\setlength{\treewidthnine}{\treewidthten}%
\setlength{\treewidthten}{\treewidtheleven}%
\setlength{\treewidtheleven}{\treewidthtwelve}%
\setlength{\treewidthtwelve}{\treewidththirteen}%
\setlength{\treewidththirteen}{\treewidthfourteen}%
\setlength{\treewidthfourteen}{\treewidthfifteen}%
\setlength{\treewidthfifteen}{\treewidthsixteen}%
\setlength{\treewidthsixteen}{\treewidthseventeen}%
\setlength{\treewidthseventeen}{\treewidtheighteen}%
\setlength{\treewidtheighteen}{\treewidthnineteen}%
\setlength{\treewidthnineteen}{\treewidthtwenty}%
\sbox{\treeboxtwo}{\usebox{\treeboxthree}}%
\sbox{\treeboxthree}{\usebox{\treeboxfour}}%
\sbox{\treeboxfour}{\usebox{\treeboxfive}}%
\sbox{\treeboxfive}{\usebox{\treeboxsix}}%
\sbox{\treeboxsix}{\usebox{\treeboxseven}}%
\sbox{\treeboxseven}{\usebox{\treeboxeight}}%
\sbox{\treeboxeight}{\usebox{\treeboxnine}}%
\sbox{\treeboxnine}{\usebox{\treeboxten}}%
\sbox{\treeboxten}{\usebox{\treeboxeleven}}%
\sbox{\treeboxeleven}{\usebox{\treeboxtwelve}}%
\sbox{\treeboxtwelve}{\usebox{\treeboxthirteen}}%
\sbox{\treeboxthirteen}{\usebox{\treeboxfourteen}}%
\sbox{\treeboxfourteen}{\usebox{\treeboxfifteen}}%
\sbox{\treeboxfifteen}{\usebox{\treeboxsixteen}}%
\sbox{\treeboxsixteen}{\usebox{\treeboxseventeen}}%
\sbox{\treeboxseventeen}{\usebox{\treeboxeighteen}}%
\sbox{\treeboxeighteen}{\usebox{\treeboxnineteen}}%
\sbox{\treeboxnineteen}{\usebox{\treeboxtwenty}}}
\newcommand{\leaf}[1]{%
\ifnum\value{treecount}=20\typeout{QobiTeX warning---Tree stack overflow}\fi%
\addtocounter{treecount}{1}%
\sbox{\treeboxtwenty}{\usebox{\treeboxnineteen}}%
\sbox{\treeboxnineteen}{\usebox{\treeboxeighteen}}%
\sbox{\treeboxeighteen}{\usebox{\treeboxseventeen}}%
\sbox{\treeboxseventeen}{\usebox{\treeboxsixteen}}%
\sbox{\treeboxsixteen}{\usebox{\treeboxfifteen}}%
\sbox{\treeboxfifteen}{\usebox{\treeboxfourteen}}%
\sbox{\treeboxfourteen}{\usebox{\treeboxthirteen}}%
\sbox{\treeboxthirteen}{\usebox{\treeboxtwelve}}%
\sbox{\treeboxtwelve}{\usebox{\treeboxeleven}}%
\sbox{\treeboxeleven}{\usebox{\treeboxten}}%
\sbox{\treeboxten}{\usebox{\treeboxnine}}%
\sbox{\treeboxnine}{\usebox{\treeboxeight}}%
\sbox{\treeboxeight}{\usebox{\treeboxseven}}%
\sbox{\treeboxseven}{\usebox{\treeboxsix}}%
\sbox{\treeboxsix}{\usebox{\treeboxfive}}%
\sbox{\treeboxfive}{\usebox{\treeboxfour}}%
\sbox{\treeboxfour}{\usebox{\treeboxthree}}%
\sbox{\treeboxthree}{\usebox{\treeboxtwo}}%
\sbox{\treeboxtwo}{\usebox{\treeboxone}}%
\sbox{\treeboxone}{\ontop{#1}}%
\sbox{\treeboxone}{\raisebox{-\ht\treeboxone}{\usebox{\treeboxone}}}%
\setlength{\treeoffsettwenty}{\treeoffsetnineteen}%
\setlength{\treeoffsetnineteen}{\treeoffseteighteen}%
\setlength{\treeoffseteighteen}{\treeoffsetseventeen}%
\setlength{\treeoffsetseventeen}{\treeoffsetsixteen}%
\setlength{\treeoffsetsixteen}{\treeoffsetfifteen}%
\setlength{\treeoffsetfifteen}{\treeoffsetfourteen}%
\setlength{\treeoffsetfourteen}{\treeoffsetthirteen}%
\setlength{\treeoffsetthirteen}{\treeoffsettwelve}%
\setlength{\treeoffsettwelve}{\treeoffseteleven}%
\setlength{\treeoffseteleven}{\treeoffsetten}%
\setlength{\treeoffsetten}{\treeoffsetnine}%
\setlength{\treeoffsetnine}{\treeoffseteight}%
\setlength{\treeoffseteight}{\treeoffsetseven}%
\setlength{\treeoffsetseven}{\treeoffsetsix}%
\setlength{\treeoffsetsix}{\treeoffsetfive}%
\setlength{\treeoffsetfive}{\treeoffsetfour}%
\setlength{\treeoffsetfour}{\treeoffsetthree}%
\setlength{\treeoffsetthree}{\treeoffsettwo}%
\setlength{\treeoffsettwo}{\treeoffsetone}%
\setlength{\treeoffsetone}{0.5\wd\treeboxone}%
\setlength{\treeshifttwenty}{\treeshiftnineteen}%
\setlength{\treeshiftnineteen}{\treeshifteighteen}%
\setlength{\treeshifteighteen}{\treeshiftseventeen}%
\setlength{\treeshiftseventeen}{\treeshiftsixteen}%
\setlength{\treeshiftsixteen}{\treeshiftfifteen}%
\setlength{\treeshiftfifteen}{\treeshiftfourteen}%
\setlength{\treeshiftfourteen}{\treeshiftthirteen}%
\setlength{\treeshiftthirteen}{\treeshifttwelve}%
\setlength{\treeshifttwelve}{\treeshifteleven}%
\setlength{\treeshifteleven}{\treeshiftten}%
\setlength{\treeshiftten}{\treeshiftnine}%
\setlength{\treeshiftnine}{\treeshifteight}%
\setlength{\treeshifteight}{\treeshiftseven}%
\setlength{\treeshiftseven}{\treeshiftsix}%
\setlength{\treeshiftsix}{\treeshiftfive}%
\setlength{\treeshiftfive}{\treeshiftfour}%
\setlength{\treeshiftfour}{\treeshiftthree}%
\setlength{\treeshiftthree}{\treeshifttwo}%
\setlength{\treeshifttwo}{\treeshiftone}%
\setlength{\treeshiftone}{0pt}%
\setlength{\treewidthtwenty}{\treewidthnineteen}%
\setlength{\treewidthnineteen}{\treewidtheighteen}%
\setlength{\treewidtheighteen}{\treewidthseventeen}%
\setlength{\treewidthseventeen}{\treewidthsixteen}%
\setlength{\treewidthsixteen}{\treewidthfifteen}%
\setlength{\treewidthfifteen}{\treewidthfourteen}%
\setlength{\treewidthfourteen}{\treewidththirteen}%
\setlength{\treewidththirteen}{\treewidthtwelve}%
\setlength{\treewidthtwelve}{\treewidtheleven}%
\setlength{\treewidtheleven}{\treewidthten}%
\setlength{\treewidthten}{\treewidthnine}%
\setlength{\treewidthnine}{\treewidtheight}%
\setlength{\treewidtheight}{\treewidthseven}%
\setlength{\treewidthseven}{\treewidthsix}%
\setlength{\treewidthsix}{\treewidthfive}%
\setlength{\treewidthfive}{\treewidthfour}%
\setlength{\treewidthfour}{\treewidththree}%
\setlength{\treewidththree}{\treewidthtwo}%
\setlength{\treewidthtwo}{\treewidthone}%
\setlength{\treewidthone}{\wd\treeboxone}}
\newcommand{\branch}[2]{%
\setcounter{branchcount}{#1}%
\ifnum\value{branchcount}=1\sbox{\parentbox}{\ontop{#2}}%
\setlength{\parentoffset}{\treeoffsetone}%
\addtolength{\parentoffset}{-0.5\wd\parentbox}%
\setlength{\daughteroffset}{0in}%
\ifdim\parentoffset<0in%
\setlength{\daughteroffset}{-\parentoffset}%
\setlength{\parentoffset}{0in}\fi%
\setlength{\parentwidth}{\parentoffset}%
\addtolength{\parentwidth}{\wd\parentbox}%
\setlength{\treeoffset}{\daughteroffset}%
\addtolength{\treeoffset}{\treeoffsetone}%
\setlength{\treewidth}{\wd\treeboxone}%
\addtolength{\treewidth}{\daughteroffset}%
\ifdim\treewidth<\parentwidth\setlength{\treewidth}{\parentwidth}\fi%
\sbox{\treebox}{\begin{minipage}{\treewidth}%
\begin{flushleft}%
\hspace*{\parentoffset}\usebox{\parentbox}\\
{\setlength{\unitlength}{2ex}%
\hspace*{\treeoffset}\begin{picture}(0,1)%
\put(0,0){\line(0,1){1}}%
\end{picture}}\\
\vspace{-\baselineskip}
\hspace*{\daughteroffset}%
\raisebox{-\ht\treeboxone}{\usebox{\treeboxone}}%
\end{flushleft}%
\end{minipage}}%
\setlength{\treeoffsetone}{\parentoffset}%
\addtolength{\treeoffsetone}{0.5\wd\parentbox}%
\setlength{\treeshiftone}{0pt}%
\setlength{\treewidthone}{\treewidth}%
\sbox{\treeboxone}{\usebox{\treebox}}%
\else\ifnum\value{branchcount}=2\sbox{\parentbox}{\ontop{#2}}%
\setlength{\branchwidthone}{\treewidthtwo}%
\addtolength{\branchwidthone}{\treeoffsetone}%
\addtolength{\branchwidthone}{-\treeshiftone}%
\addtolength{\branchwidthone}{-\treeoffsettwo}%
\setlength{\branchwidth}{\branchwidthone}%
\setlength{\daughteroffsetone}{\branchwidth}%
\addtolength{\daughteroffsetone}{-\branchwidthone}%
\addtolength{\daughteroffsetone}{-\treeshiftone}%
\setlength{\parentoffset}{-0.5\wd\parentbox}%
\addtolength{\parentoffset}{\treeoffsettwo}%
\addtolength{\parentoffset}{0.5\branchwidth}%
\setlength{\daughteroffset}{0in}%
\ifdim\parentoffset<0in%
\setlength{\daughteroffset}{-\parentoffset}%
\setlength{\parentoffset}{0in}\fi%
\setlength{\parentwidth}{\parentoffset}%
\addtolength{\parentwidth}{\wd\parentbox}%
\setlength{\treeoffset}{\daughteroffset}%
\addtolength{\treeoffset}{\treeoffsettwo}%
\setlength{\treewidth}{\wd\treeboxone}%
\addtolength{\treewidth}{\daughteroffsetone}%
\addtolength{\treewidth}{\treewidthtwo}%
\addtolength{\treewidth}{\daughteroffset}%
\ifdim\treewidth<\parentwidth\setlength{\treewidth}{\parentwidth}\fi%
\sbox{\treebox}{\begin{minipage}{\treewidth}%
\begin{flushleft}%
\hspace*{\parentoffset}\usebox{\parentbox}\\
{\setlength{\unitlength}{0.5\branchwidth}%
\hspace*{\treeoffset}\begin{picture}(2,0.5)%
\put(0,0){\line(2,1){1}}%
\put(2,0){\line(-2,1){1}}%
\end{picture}}\\
\vspace{-\baselineskip}
\hspace*{\daughteroffset}%
\makebox[\treewidthtwo][l]%
{\raisebox{-\ht\treeboxtwo}{\usebox{\treeboxtwo}}}%
\hspace*{\daughteroffsetone}%
\raisebox{-\ht\treeboxone}{\usebox{\treeboxone}}%
\end{flushleft}%
\end{minipage}}%
\setlength{\treeoffsetone}{\parentoffset}%
\addtolength{\treeoffsetone}{0.5\wd\parentbox}%
\setlength{\treeshiftone}{0pt}%
\setlength{\treewidthone}{\treewidth}%
\sbox{\treeboxone}{\usebox{\treebox}}\poptree%
\else\ifnum\value{branchcount}=3\sbox{\parentbox}{\ontop{#2}}%
\setlength{\branchwidthone}{\treewidthtwo}%
\addtolength{\branchwidthone}{\treeoffsetone}%
\addtolength{\branchwidthone}{-\treeshiftone}%
\addtolength{\branchwidthone}{-\treeoffsettwo}%
\setlength{\branchwidthtwo}{\treewidththree}%
\addtolength{\branchwidthtwo}{\treeoffsettwo}%
\addtolength{\branchwidthtwo}{-\treeshifttwo}%
\addtolength{\branchwidthtwo}{-\treeoffsetthree}%
\setlength{\branchwidth}{\branchwidthone}%
\ifdim\branchwidthtwo>\branchwidth%
\setlength{\branchwidth}{\branchwidthtwo}\fi%
\setlength{\daughteroffsetone}{\branchwidth}%
\addtolength{\daughteroffsetone}{-\branchwidthone}%
\addtolength{\daughteroffsetone}{-\treeshiftone}%
\setlength{\daughteroffsettwo}{\branchwidth}%
\addtolength{\daughteroffsettwo}{-\branchwidthtwo}%
\addtolength{\daughteroffsettwo}{-\treeshifttwo}%
\setlength{\parentoffset}{-0.5\wd\parentbox}%
\addtolength{\parentoffset}{\treeoffsetthree}%
\addtolength{\parentoffset}{\branchwidth}%
\setlength{\daughteroffset}{0in}%
\ifdim\parentoffset<0in%
\setlength{\daughteroffset}{-\parentoffset}%
\setlength{\parentoffset}{0in}\fi%
\setlength{\parentwidth}{\parentoffset}%
\addtolength{\parentwidth}{\wd\parentbox}%
\setlength{\treeoffset}{\daughteroffset}%
\addtolength{\treeoffset}{\treeoffsetthree}%
\setlength{\treewidth}{\wd\treeboxone}%
\addtolength{\treewidth}{\daughteroffsetone}%
\addtolength{\treewidth}{\treewidthtwo}%
\addtolength{\treewidth}{\daughteroffsettwo}%
\addtolength{\treewidth}{\treewidththree}%
\addtolength{\treewidth}{\daughteroffset}%
\ifdim\treewidth<\parentwidth\setlength{\treewidth}{\parentwidth}\fi%
\sbox{\treebox}{\begin{minipage}{\treewidth}%
\begin{flushleft}%
\hspace*{\parentoffset}\usebox{\parentbox}\\
{\setlength{\unitlength}{0.5\branchwidth}%
\hspace*{\treeoffset}\begin{picture}(4,1)%
\put(0,0){\line(2,1){2}}%
\put(2,0){\line(0,1){1}}%
\put(4,0){\line(-2,1){2}}%
\end{picture}}\\
\vspace{-\baselineskip}
\hspace*{\daughteroffset}%
\makebox[\treewidththree][l]%
{\raisebox{-\ht\treeboxthree}{\usebox{\treeboxthree}}}%
\hspace*{\daughteroffsettwo}%
\makebox[\treewidthtwo][l]%
{\raisebox{-\ht\treeboxtwo}{\usebox{\treeboxtwo}}}%
\hspace*{\daughteroffsetone}%
\raisebox{-\ht\treeboxone}{\usebox{\treeboxone}}%
\end{flushleft}%
\end{minipage}}%
\setlength{\treeoffsetone}{\parentoffset}%
\addtolength{\treeoffsetone}{0.5\wd\parentbox}%
\setlength{\treeshiftone}{0pt}%
\setlength{\treewidthone}{\treewidth}%
\sbox{\treeboxone}{\usebox{\treebox}}\poptree\poptree%
\else\ifnum\value{branchcount}=4\sbox{\parentbox}{\ontop{#2}}%
\setlength{\branchwidthone}{\treewidthtwo}%
\addtolength{\branchwidthone}{\treeoffsetone}%
\addtolength{\branchwidthone}{-\treeshiftone}%
\addtolength{\branchwidthone}{-\treeoffsettwo}%
\setlength{\branchwidthtwo}{\treewidththree}%
\addtolength{\branchwidthtwo}{\treeoffsettwo}%
\addtolength{\branchwidthtwo}{-\treeshifttwo}%
\addtolength{\branchwidthtwo}{-\treeoffsetthree}%
\setlength{\branchwidththree}{\treewidthfour}%
\addtolength{\branchwidththree}{\treeoffsetthree}%
\addtolength{\branchwidththree}{-\treeshiftthree}%
\addtolength{\branchwidththree}{-\treeoffsetfour}%
\setlength{\branchwidth}{\branchwidthone}%
\ifdim\branchwidthtwo>\branchwidth%
\setlength{\branchwidth}{\branchwidthtwo}\fi%
\ifdim\branchwidththree>\branchwidth%
\setlength{\branchwidth}{\branchwidththree}\fi%
\setlength{\daughteroffsetone}{\branchwidth}%
\addtolength{\daughteroffsetone}{-\branchwidthone}%
\addtolength{\daughteroffsetone}{-\treeshiftone}%
\setlength{\daughteroffsettwo}{\branchwidth}%
\addtolength{\daughteroffsettwo}{-\branchwidthtwo}%
\addtolength{\daughteroffsettwo}{-\treeshifttwo}%
\setlength{\daughteroffsetthree}{\branchwidth}%
\addtolength{\daughteroffsetthree}{-\branchwidththree}%
\addtolength{\daughteroffsetthree}{-\treeshiftthree}%
\setlength{\parentoffset}{-0.5\wd\parentbox}%
\addtolength{\parentoffset}{\treeoffsetfour}%
\addtolength{\parentoffset}{1.5\branchwidth}%
\setlength{\daughteroffset}{0in}%
\ifdim\parentoffset<0in%
\setlength{\daughteroffset}{-\parentoffset}%
\setlength{\parentoffset}{0in}\fi%
\setlength{\parentwidth}{\parentoffset}%
\addtolength{\parentwidth}{\wd\parentbox}%
\setlength{\treeoffset}{\daughteroffset}%
\addtolength{\treeoffset}{\treeoffsetfour}%
\setlength{\treewidth}{\wd\treeboxone}%
\addtolength{\treewidth}{\daughteroffsetone}%
\addtolength{\treewidth}{\treewidthtwo}%
\addtolength{\treewidth}{\daughteroffsettwo}%
\addtolength{\treewidth}{\treewidththree}%
\addtolength{\treewidth}{\daughteroffsetthree}%
\addtolength{\treewidth}{\treewidthfour}%
\addtolength{\treewidth}{\daughteroffset}%
\ifdim\treewidth<\parentwidth\setlength{\treewidth}{\parentwidth}\fi%
\sbox{\treebox}{\begin{minipage}{\treewidth}%
\begin{flushleft}%
\hspace*{\parentoffset}\usebox{\parentbox}\\
{\setlength{\unitlength}{0.5\branchwidth}%
\hspace*{\treeoffset}\begin{picture}(6,1)%
\put(0,0){\line(3,1){3}}%
\put(2,0){\line(1,1){1}}%
\put(4,0){\line(-1,1){1}}%
\put(6,0){\line(-3,1){3}}%
\end{picture}}\\
\vspace{-\baselineskip}
\hspace*{\daughteroffset}%
\makebox[\treewidthfour][l]%
{\raisebox{-\ht\treeboxfour}{\usebox{\treeboxfour}}}%
\hspace*{\daughteroffsetthree}%
\makebox[\treewidththree][l]%
{\raisebox{-\ht\treeboxthree}{\usebox{\treeboxthree}}}%
\hspace*{\daughteroffsettwo}%
\makebox[\treewidthtwo][l]%
{\raisebox{-\ht\treeboxtwo}{\usebox{\treeboxtwo}}}%
\hspace*{\daughteroffsetone}%
\raisebox{-\ht\treeboxone}{\usebox{\treeboxone}}%
\end{flushleft}%
\end{minipage}}%
\setlength{\treeoffsetone}{\parentoffset}%
\addtolength{\treeoffsetone}{0.5\wd\parentbox}%
\setlength{\treeshiftone}{0pt}%
\setlength{\treewidthone}{\treewidth}%
\sbox{\treeboxone}{\usebox{\treebox}}\poptree\poptree\poptree%
\else\ifnum\value{branchcount}=5\sbox{\parentbox}{\ontop{#2}}%
\setlength{\branchwidthone}{\treewidthtwo}%
\addtolength{\branchwidthone}{\treeoffsetone}%
\addtolength{\branchwidthone}{-\treeshiftone}%
\addtolength{\branchwidthone}{-\treeoffsettwo}%
\setlength{\branchwidthtwo}{\treewidththree}%
\addtolength{\branchwidthtwo}{\treeoffsettwo}%
\addtolength{\branchwidthtwo}{-\treeshifttwo}%
\addtolength{\branchwidthtwo}{-\treeoffsetthree}%
\setlength{\branchwidththree}{\treewidthfour}%
\addtolength{\branchwidththree}{\treeoffsetthree}%
\addtolength{\branchwidththree}{-\treeshiftthree}%
\addtolength{\branchwidththree}{-\treeoffsetfour}%
\setlength{\branchwidthfour}{\treewidthfive}%
\addtolength{\branchwidthfour}{\treeoffsetfour}%
\addtolength{\branchwidthfour}{-\treeshiftfour}%
\addtolength{\branchwidthfour}{-\treeoffsetfive}%
\setlength{\branchwidth}{\branchwidthone}%
\ifdim\branchwidthtwo>\branchwidth%
\setlength{\branchwidth}{\branchwidthtwo}\fi%
\ifdim\branchwidththree>\branchwidth%
\setlength{\branchwidth}{\branchwidththree}\fi%
\ifdim\branchwidthfour>\branchwidth%
\setlength{\branchwidth}{\branchwidthfour}\fi%
\setlength{\daughteroffsetone}{\branchwidth}%
\addtolength{\daughteroffsetone}{-\branchwidthone}%
\addtolength{\daughteroffsetone}{-\treeshiftone}%
\setlength{\daughteroffsettwo}{\branchwidth}%
\addtolength{\daughteroffsettwo}{-\branchwidthtwo}%
\addtolength{\daughteroffsettwo}{-\treeshifttwo}%
\setlength{\daughteroffsetthree}{\branchwidth}%
\addtolength{\daughteroffsetthree}{-\branchwidththree}%
\addtolength{\daughteroffsetthree}{-\treeshiftthree}%
\setlength{\daughteroffsetfour}{\branchwidth}%
\addtolength{\daughteroffsetfour}{-\branchwidthfour}%
\addtolength{\daughteroffsetfour}{-\treeshiftfour}%
\setlength{\parentoffset}{-0.5\wd\parentbox}%
\addtolength{\parentoffset}{\treeoffsetfive}%
\addtolength{\parentoffset}{2\branchwidth}%
\setlength{\daughteroffset}{0in}%
\ifdim\parentoffset<0in%
\setlength{\daughteroffset}{-\parentoffset}%
\setlength{\parentoffset}{0in}\fi%
\setlength{\parentwidth}{\parentoffset}%
\addtolength{\parentwidth}{\wd\parentbox}%
\setlength{\treeoffset}{\daughteroffset}%
\addtolength{\treeoffset}{\treeoffsetfive}%
\setlength{\treewidth}{\wd\treeboxone}%
\addtolength{\treewidth}{\daughteroffsetone}%
\addtolength{\treewidth}{\treewidthtwo}%
\addtolength{\treewidth}{\daughteroffsettwo}%
\addtolength{\treewidth}{\treewidththree}%
\addtolength{\treewidth}{\daughteroffsetthree}%
\addtolength{\treewidth}{\treewidthfour}%
\addtolength{\treewidth}{\daughteroffsetfour}%
\addtolength{\treewidth}{\treewidthfive}%
\addtolength{\treewidth}{\daughteroffset}%
\ifdim\treewidth<\parentwidth\setlength{\treewidth}{\parentwidth}\fi%
\sbox{\treebox}{\begin{minipage}{\treewidth}%
\begin{flushleft}%
\hspace*{\parentoffset}\usebox{\parentbox}\\
{\setlength{\unitlength}{0.5\branchwidth}%
\hspace*{\treeoffset}\begin{picture}(8,1)%
\put(0,0){\line(4,1){4}}%
\put(2,0){\line(2,1){2}}%
\put(4,0){\line(0,1){1}}%
\put(6,0){\line(-2,1){2}}%
\put(8,0){\line(-4,1){4}}%
\end{picture}}\\
\vspace{-\baselineskip}
\hspace*{\daughteroffset}%
\makebox[\treewidthfive][l]%
{\raisebox{-\ht\treeboxfour}{\usebox{\treeboxfive}}}%
\hspace*{\daughteroffsetfour}%
\makebox[\treewidthfour][l]%
{\raisebox{-\ht\treeboxfour}{\usebox{\treeboxfour}}}%
\hspace*{\daughteroffsetthree}%
\makebox[\treewidththree][l]%
{\raisebox{-\ht\treeboxthree}{\usebox{\treeboxthree}}}%
\hspace*{\daughteroffsettwo}%
\makebox[\treewidthtwo][l]%
{\raisebox{-\ht\treeboxtwo}{\usebox{\treeboxtwo}}}%
\hspace*{\daughteroffsetone}%
\raisebox{-\ht\treeboxone}{\usebox{\treeboxone}}%
\end{flushleft}%
\end{minipage}}%
\setlength{\treeoffsetone}{\parentoffset}%
\addtolength{\treeoffsetone}{0.5\wd\parentbox}%
\setlength{\treeshiftone}{0pt}%
\setlength{\treewidthone}{\treewidth}%
\sbox{\treeboxone}{\usebox{\treebox}}\poptree\poptree\poptree\poptree%
\else\typeout{QobiTeX warning--- Can't handle #1 branching}\fi\fi\fi\fi\fi}
\newcommand{\tree}{%
\usebox{\treeboxone}
\setlength{\treeoffsetone}{\treeoffsettwo}%
\sbox{\treeboxone}{\usebox{\treeboxtwo}}%
\poptree}
\makeatother

\makeatletter

\newcounter{enums}

\newdimen\widelabel
\newlength{\exskip}
\def\enumsentence{\@ifnextchar[{\@enumsentence}
{\refstepcounter{enums}\@enumsentence[(\theenums)]}}

\long\def\@enumsentence[#1]#2{\begin{list}{}{%
\advance\leftmargin by\widelabel \advance\labelwidth by \widelabel%
\setlength{\labelsep}{\exskip} \advance\leftmargin by 1.1em}
\item[#1] #2
\end{list}}

\newcounter{tempcnt}

\def\@item[#1]{\if@noparitem \@donoparitem
  \else \if@inlabel \indent \par \fi
         \ifhmode \unskip\unskip \par \fi 
         \if@newlist \if@nobreak \@nbitem \else
                        \addpenalty\@beginparpenalty
                        \addvspace\@topsep \addvspace{-\parskip}\fi
           \else \addpenalty\@itempenalty \addvspace\itemsep 
          \fi 
    \global\@inlabeltrue 
\fi
\everypar{\global\@minipagefalse\global\@newlistfalse 
          \if@inlabel\global\@inlabelfalse \hskip -\parindent \box\@labels
             \penalty\z@ \fi
          \everypar{}}\global\@nobreakfalse
\if@noitemarg \@noitemargfalse \if@nmbrlist \refstepcounter{\@listctr}\fi \fi
\setbox\@tempboxa\hbox{\makelabel{#1}}%
\global\setbox\@labels
 \hbox{\unhbox\@labels \hskip \itemindent
       \hskip -\labelwidth \hskip -\labelsep 
       \ifdim \wd\@tempboxa >\labelwidth 
                \box\@tempboxa
          \else \hbox to\labelwidth {\unhbox\@tempboxa}\fi
       \hskip \labelsep}\ignorespaces}

\newcounter{enumsi}

\newdimen\eeindent
\def\@mklab#1{\hfil#1}
\def\enummklab#1{\hfil(\eelabel)\hbox to \eeindent{\hfil#1}}
\def\enummakelabel#1{\enummklab{#1}\global\let\makelabel=\@mklab}
\def\toplabel#1{{\edef\@currentlabel{\p@enums\theenums}\label{#1}}}

\def\eenumsentence{\@ifnextchar[{\@eenumsentence}
{\refstepcounter{enums}\@eenumsentence[\theenums]}}

\long\def\@eenumsentence[#1]#2{\def\eelabel{#1}\let\holdlabel\makelabel%
\begin{list}{\alph{enumsi}.}{%
\itemsep=0pt \parsep=0pt
\usecounter{enumsi}%
\advance\leftmargin by \eeindent \advance\leftmargin by \widelabel%
\advance\labelwidth by \eeindent \advance\labelwidth by \widelabel%
\let\makelabel=\enummakelabel%
}
#2
\end{list}\let\makelabel\holdlabel}

\makeatother

\title{Automatically Creating Bilingual Lexicons for Machine
Translation from Bilingual Text}

\author{Davide Turcato\\
\begin{tabular}{cc}
Natural Language Lab		& TCC Communications\\
School of Computing Science	& 100-6722 Oldfield Road\\
Simon Fraser University		& Victoria, BC\\
Burnaby, BC, V5A 1S6		& V8M 2A3\\
Canada				& Canada\\
{\tt turk@cs.sfu.ca}		& {\tt turk@tcc.bc.ca}
\end{tabular}
}

\begin{document}

\maketitle

\abstract{A method is presented for automatically augmenting the
bilingual lexicon of an existing Machine Translation system, by
extracting bilingual entries from aligned bilingual text. The proposed
method only relies on the resources already available in the MT system
itself. It is based on the use of bilingual lexical templates to match
the terminal symbols in the parses of the aligned sentences.}

\section{Introduction}

A novel approach to automatically building bilingual lexicons is
presented here. The term {\em bilingual lexicon} denotes a collection
of complex equivalences as used in Machine Translation (MT) transfer
lexicons, not just word equivalences. In addition to words, such
lexicons involve syntactic and semantic descriptions and means to
perform a correct transfer between the two sides of a bilingual
lexical entry.

A symbolic, rule-based approach of the {\em parse-parse-match} kind is
proposed. The core idea is to use the resources of bidirectional
transfer MT systems for this purpose, taking advantage of their
features to convert them to a novel use. In addition to having them
use their bilingual lexicons to produce translations, it is proposed
to have them use translations to produce bilingual lexicons. Although
other uses might be conceived, the most appropriate use is to have an
MT system automatically augment its own bilingual lexicon from a small
initial sample.

The core of the described approach consists of using a set of
bilingual lexical templates in matching the parses of two aligned
sentences and in turning the lexical equivalences thus established
into new bilingual lexical entries.

\section{Theoretical framework}

The basic requirement that an MT system should meet for the present
purpose is to be {\em bidirectional}. Bidirectionality is required in
order to ensure that both source and target grammars can be used for
parsing and that transfer can be done in both directions. More
precisely, what is relevant is that the input and output to transfer
be the same kind of structure.

Moreover, the proposed method is most productive with a lexicalist MT
system \cite{Whitelock:SB}. The proposed application is concerned with
producing bilingual lexical knowledge and this sort of knowledge is
the only type of bilingual knowledge required by lexicalist
systems. Nevertheless, it is also conceivable that the present
approach can be used with a non-lexicalist transfer system, as long as
the system is bidirectional. In this case, only the lexical portion of
the bilingual knowledge can be automatically produced, assuming that
the structural transfer portion is already in place. In the rest of
this paper, a lexicalist MT system will be assumed and referred
to. For the specific implementation described here and all the
examples, we will refer to an existing lexicalist English-Spanish MT
system \cite{Popowich:TMI}.

The main feature of a lexicalist MT system is that it performs no
structural transfer. Transfer is a mapping between a bag of lexical
items used in parsing (the {\em source bag}) and a corresponding bag
of target lexical items (the {\em target bag}), to be used in
generation. The source bag actually contains more information than the
corresponding bag of lexical items before parsing. Its elements get
enriched with additional information instantiated during the parsing
process. Information of fundamental importance included therein is a
system of indices that express dependencies among lexical items. Such
dependencies are transferred to the target bag and used to constrain
generation. The task of generation is to find an order in which the
lexical items can be successfully parsed.

\section{Bilingual templates}

A {\em bilingual template} is a bilingual entry in which
words are left unspecified. E.g.:

\enumsentence{\label{bt} \tt \_ :: (L,@count\_noun(A))
$\leftrightarrow$\\ \_ :: (R,@noun(A))
$\backslash\backslash$trans\_noun(L,R).}

Here, a `{\tt ::}' operator connects a word (a variable, in a
template) to a description, `{\tt $\leftrightarrow$}' connects the
left and right sides of the entry, `{\tt $\backslash\backslash$}'
introduces a {\em transfer macro}, which takes two descriptions as
arguments and performs some additional transfer
\cite{Turcato:RANLP97}. Descriptions are mainly expressed by macros,
introduced by a `{\tt @}' operator. The macro arguments are indices,
as used in lexicalist transfer.

Templates have been widely used in MT \cite{Buschbeck:RANLP97},
particularly in the Example-Based Machine Translation (EBMT) framework
(\newcite{Kaji:COLING92}, \newcite{Guvenir:CSCSI96}). However, in
EBMT, templates are most often used to model sentence-level
correspondences, rather then lexical equivalences. Consequently, in
EBMT the relation between lexical equivalences and templates is the
reverse of what is being proposed here. In EBMT, lexical equivalences
are assumed and (sentential) templates are inferred from them. In the
present framework, sentential correspondences (in the form of possible
combinations of lexical templates) are assumed and lexical
equivalences are inferred from them.

In a lexicalist approach, the notion of bilingual lexical entry, and
thus that of bilingual template, must be intended broadly. Multiword
entries can exist. They can express dependencies among lexical items,
thus being suitable for expressing phrasal equivalences. In brief,
bilingual lexical entries can exhaustively cover all the bilingual
information needed in transfer.

In a lexicalist MT system, transfer is accomplished by finding a bag
of bilingual entries partitioning the source bag. The source side of
each entry (in the rest of this paper: the left hand side) corresponds
to a cell of the partition. The union of the target sides of the
entries constitutes the target bag. E.g.:

\eenumsentence{ \label{bilex-equiv}

\item \label{Source-bag} Source bag:

\{{\em Sw}$_{1}$::{\em Sd}$_{1}$, {\em Sw}$_{2}$::{\em Sd}$_{2}$, {\em
Sw}$_{3}$::{\em Sd}$_{3}$\}

\medskip

\item \label{Bilingual-entries} Bilingual entries:

\{{\em Sw}$_{1}$::{\em Sd}$_{1}$ \& {\em Sw}$_{3}$::{\em Sd}$_{3}$
$\leftrightarrow$\\ {\em Tw}$_{1}$::{\em Td}$_{1}$ \& {\em
Tw}$_{2}$::{\em Td}$_{2}$,

\medskip

{\em Sw}$_{2}$::{\em Sd}$_{2}$ $\leftrightarrow$\\ {\em Tw}$_{3}$::{\em
Td}$_{3}$ \& {\em Tw}$_{4}$::{\em Td}$_{4}$\}

\medskip

\item \label{Target-bag} Target bag:

\{{\em Tw}$_{1}$::{\em Td}$_{1}$, {\em Tw}$_{2}$::{\em Td}$_{2}$, {\em
Tw}$_{3}$::{\em Td}$_{3}$,\\
{\em Tw}$_{4}$::{\em Td}$_{4}$\}

}

\noindent where each {\em Sw}$_{i}$::{\em Sd}$_{i}$ and {\em
Tw}$_{i}$::{\em Td}$_{i}$ are, respectively, a source and target
$<${\em Word},{\em Description}$>$ pair. In addition, the bilingual
entries must satisfy the constraints expressed by indices in the
source and target bags. The same information can be used to find
(\ref{Bilingual-entries}), given (\ref{Source-bag}) and
(\ref{Target-bag}).

Any bilingual lexicon is partitioned by a set of templates. The
entries in each equivalence class only differ by their words. A
bilingual lexical entry can thus be viewed as a triple $<${\em
Sw,Tw,T}$>$, where {\em Sw} is a list of source words, {\em Tw} a list
of target words, and {\em T} a template. A set of such bilingual
templates can be intuitively regarded as a `transfer grammar'. A
grammar defines all the possible sequences of pre-terminal symbols,
i.e. all the possible types of sentences. Analogously, a set of
bilingual templates defines all the possible translational
equivalences between bags of pre-terminal symbols, i.e. all the
possible equivalences between types of sentences.

Using this intuition, the possibility is explored of analyzing a pair
of such bags by means of a database of bilingual templates, to find a
bag of templates that correctly accounts for the translational
equivalence of the two bags, without resorting to any information
about words. In the example (\ref{bilex-equiv}), the following bag of
templates would be the requested solution:

\enumsentence{\label{Bilingual-templates}

\{\_::{\em Sd}$_{1}$ \& \_::{\em Sd}$_{3}$ $\leftrightarrow$ \_::{\em
Td}$_{1}$ \& \_::{\em Td}$_{2}$,

\medskip

\_::{\em Sd}$_{2}$ $\leftrightarrow$ \_::{\em Td}$_{3}$ \& \_::{\em
Td}$_{4}$\} }

Equivalences between (bags of) words are automatically obtained as a
result of the process, whereas in translating they are assumed and
used to select the appropriate bilingual entries.

The whole idea is based on the assumption that a lexical item's
description and the constraints on its indices are sufficient in most
cases to uniquely identify a lexical item in a parse output
bag. Although exceptions could be found (most notably, two modifiers
of the same category modifying the same head), the idea is viable
enough to be worth exploring.

The impression might arise that it is difficult and impractical to
have a set of templates available in advance. However, there is
empirical evidence to the contrary. A count on the MT system used here
showed that a restricted number of templates covers a large portion of
a bilingual lexicon. Table \ref{coverage} shows the incremental
coverage. Although completeness is hard to obtain, a satisfactory
coverage can be achieved with a relatively small number of templates.

\begin{table}

\begin{center}

\begin{tabular}{rrr}

\hline

Templates	& Entries 	& Coverage		\\

\hline

1		& 5683		& 33.9 \%		\\
2		& 8726		& 52.1 \%		\\
3		& 10710		& 63.9 \%		\\
4		& 12336		& 73.6 \%		\\
5		& 13609		& 81.2 \%		\\
50		& 15473		& 92.3 \%		\\
500		& 16338		& 97.5 \%		\\
922		& 16760		& 100.0 \%		\\

\hline

\end{tabular}

\end{center}

\caption{\label{coverage}Incremental template coverage}

\end{table}

In the implementation described here, a set of templates was extracted
from the MT bilingual lexicon and used to bootstrap further lexical
development. The whole lexical development can be seen as an
interactive process involving a bilingual lexicon and a template
database. Templates are initially derived from the lexicon, new
entries are successively created using the templates. Iteratively, new
entries can be manually coded when the automatic procedure is lacking
appropriate templates and new templates extracted from the manually
coded entries can be added to the template database.

\section{The algorithm}

In this section the algorithm for creating bilingual lexical entries
is described, along with a sample run. The procedure was implemented
in Prolog, as was the MT system at hand. Basically, a set of lexical
entries is obtained from a pair of sentences by first parsing the
source and target sentences. The source bag is then transferred using
templates as transfer rules (plus entries for closed-class words and
possibly a pre-existing bilingual lexicon). The transfer output bag is
then unified with the target sentence parse output bag. If the
unification succeeds, the relevant information (bilingual templates
and associated words) is retrieved to build up the new bilingual
entries. Otherwise, the system backtracks into new parses and
transfers.

The main predicate {\tt make\_entries/3} matches a source and a target
sentence to produce a set of bilingual entries:

\medskip

\begin{verbatim}
make_entries(Source,Target,Entries):-
   parse_source(Source,Deriv1),
   parse_target(Target,Deriv2),
   transfer(Deriv1,Deriv3),
   get_bag(Deriv2,Bag2),
   get_bag(Deriv3,Bag3),
   match_bags(Bag2,Bag3,Bag4),
   get_bag(Deriv1,Bag1),
   make_be_info(Bag1,Bag4,Deriv3,Be),
   be_info_to_entries(Be,Entries).
\end{verbatim}

\medskip

Each {\tt Deriv}{\em n} variable points to a buffer where all the
information about a specific derivation (parse or transfer) is stored
and each {\tt Bag}{\em n} variable refers to a bag of lexical
items. Each step will be discussed in detail in the rest of the
section. A sample run will be shown for the following English-Spanish
pair of sentences:

\eenumsentence{

\label{pair}

\item {\tt the fat man kicked out the black dog.}

\item {\tt el hombre gordo ech\'{o} el perro negro.}

}

In the sample session no bilingual lexicon was used for content
words. Only a bilingual lexicon for closed class words and a set of
bilingual templates were used. Therefore, new bilingual entries were
obtained for all the content words (or phrases) in the sentences.

\subsection{Source sentence parse}

\begin{figure}
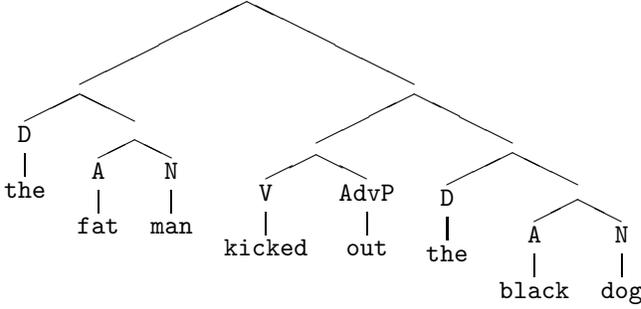

\small
{\tt
\leaf{the}
\branch{1}{D}
\leaf{fat}
\branch{1}{A}
\leaf{man}
\branch{1}{N}
\branch{2}{}
\branch{2}{}
\leaf{kicked}
\branch{1}{V}
\leaf{out}
\branch{1}{AdvP}
\branch{2}{}
\leaf{the}
\branch{1}{D}
\leaf{black}
\branch{1}{A}
\leaf{dog}
\branch{1}{N}
\branch{2}{}
\branch{2}{}
\branch{2}{}
\branch{2}{}
\tree
}
\normalsize
\caption{\label{SSPT}Source sentence parse tree.}
\end{figure}

The parse of the source sentence is performed by {\tt
parse\_source/2}. The parse tree is shown in Fig. \ref{SSPT}. Since
only lexical items are relevant for the present purposes, only
pre-terminal nodes in the tree are labeled.
 
Fig. \ref{SSPOB} shows, in succint form, the relevant information from
the source bag, i.e. the bag resulting from parsing the source
sentence. All the syntactic and semantic information has been omitted
and replaced by a category label. What is relevant here is the way the
indices are set, as a result of parsing. The words \{{\tt
the,fat,man}\} are tied together and so are \{{\tt kick,out}\} and
\{{\tt the,black,dog}\}. Moreover, the indices of `{\tt kick}' show
that its second index is tied to its subject, \{{\tt the,fat,man}\},
and its third index is tied to its object, \{{\tt the,black,dog}\}.

\begin{figure}
\begin{tabular}{llll}

{\tt Id}& {\tt Word}	& {\tt Cat}		& {\tt Indices}	\\ \hline
{\tt 1}	& {\tt the}	& {\tt determiner}	& {\tt [0]}	\\
{\tt 2}	& {\tt fat}	& {\tt adjective}	& {\tt [0]}	\\
{\tt 3}	& {\tt man}	& {\tt noun}		& {\tt [0]}	\\
{\tt 4}	& {\tt kick}	& {\tt trans\_verb}	& {\tt [10,0,9]}\\
{\tt 5}	& {\tt out}	& {\tt advparticle}	& {\tt [10]}	\\
{\tt 6}	& {\tt the}	& {\tt determiner}	& {\tt [9]}	\\
{\tt 7}	& {\tt black}	& {\tt adjective}	& {\tt [9]}	\\
{\tt 8}	& {\tt dog}	& {\tt noun}		& {\tt [9]}	\\

\end{tabular}
\caption{\label{SSPOB}Source sentence parse output bag.}
\end{figure}

\subsection{Target sentence parse}

\begin{figure}
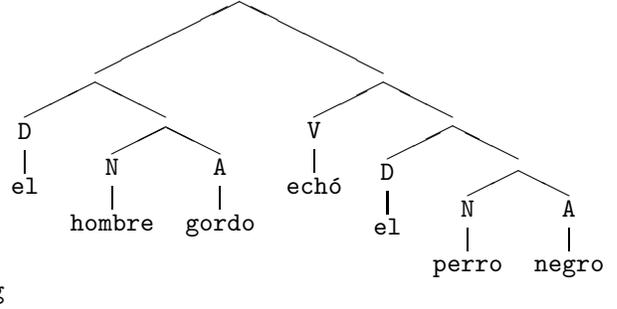

\small
{\tt
\leaf{el}
\branch{1}{D}
\leaf{hombre}
\branch{1}{N}
\leaf{gordo}
\branch{1}{A}
\branch{2}{}
\branch{2}{}
\leaf{ech\'{o}}
\branch{1}{V}
\leaf{el}
\branch{1}{D}
\leaf{perro}
\branch{1}{N}
\leaf{negro}
\branch{1}{A}
\branch{2}{}
\branch{2}{}
\branch{2}{}
\branch{2}{}
\tree
}
\normalsize
\caption{\label{TSPT}Target sentence parse tree.}
\end{figure}

The parse of the target sentence is performed by {\tt
parse\_target/2}. Fig. \ref{TSPT} and \ref{TSPOB} show, respectively,
the resulting tree and bag. In an analogous manner to what is seen in
the source sentence, \{{\tt el,hombre,gordo}\} and \{{\tt
el,perro,negro}\} are, respectively, the subject and the object of
`{\tt ech\'{o}}'.

\begin{figure}
\begin{tabular}{llll}

{\tt Id}& {\tt Word}	& {\tt Cat}	& {\tt Indices}		\\ \hline
{\tt 1}	& {\tt el}	& {\tt d}	& {\tt [0]}		\\
{\tt 2}	& {\tt hombre}	& {\tt n}	& {\tt [0]}		\\
{\tt 3}	& {\tt gordo}	& {\tt adj}	& {\tt [0]}		\\
{\tt 4}	& {\tt echar}	& {\tt v}	& {\tt [1,0,13]}	\\
{\tt 5}	& {\tt el}	& {\tt d}	& {\tt [13]}		\\
{\tt 6}	& {\tt perro}	& {\tt n}	& {\tt [13]}		\\
{\tt 7}	& {\tt negro}	& {\tt adj}	& {\tt [13]}		\\

\end{tabular}
\caption{\label{TSPOB}Target sentence parse output bag.}
\end{figure}

\subsection{Transfer}

\begin{figure}
\begin{tabular}{llll}

{\tt Id}  & {\tt Word}		    & {\tt Cat} & {\tt Indices}	\\ \hline
{\tt 2-1} & {\tt el}		    & {\tt d}   & {\tt [A]}	\\
{\tt 3-2} & {\tt word(adj/adj,1)}   & {\tt adj} & {\tt [A]}	\\
{\tt 4-3} & {\tt word(cn/n,1)}	    & {\tt n}   & {\tt [A]}	\\
{\tt 1-4} & {\tt word(tv+adv/tv,1)} & {\tt v}   & {\tt [B,A,I]}	\\
{\tt 5-6} & {\tt el}		    & {\tt d}   & {\tt [I]}	\\
{\tt 6-7} & {\tt word(adj/adj,1)}   & {\tt adj} & {\tt [I]}	\\
{\tt 7-8} & {\tt word(cn/n,1)}	    & {\tt n}   & {\tt [I]}	\\

\end{tabular}
\caption{\label{TOB}Transfer output bag.}
\end{figure}

The result of parsing the source sentence is used by {\tt transfer/2}
to create a translationally equivalent target bag. Fig. \ref{TOB}
shows the result. Transfer is performed by consulting a bilingual
lexicon, which, in the present case, contained entries for closed
class words (e.g. an entry mapping `{\tt the}' to `{\tt el}') and
templates for content words. The templates relevant to our example are
the following:

\eenumsentence{

\item{\tt 
\_ ::@adj(A)\\
$\leftrightarrow$ 'word(adj/adj,1)' ::@adj(A).\\
}

\item{\tt 
\_ ::(L,@count\_noun(A))\\
$\leftrightarrow$ 'word(cn/n,1)' ::(R,@noun(A))\\
$\backslash\backslash$trans\_noun(L,R).\\
}

\item{\tt 
\_ ::(L,@trans\_verb(A,B,C))\\
\& \_ ::@advparticle(A)\\
$\leftrightarrow$\\
'word(tv+adv/tv,1)' :: (R,@verb\_acc(A,B,C))\\
$\backslash\backslash$trans\_verb(L,R).
}

}

Bilingual templates are simply bilingual entries with words replaced
by variables. Actually, on the target side, words are replaced by
labels of the form {\tt word(Ti,Position)}, where {\tt Ti} is a
template identifier and {\tt Position} identifies the position of the
item in the right hand side of the template. Thus, a label {\tt
word(adj/adj,1)} identifies the first word on the right hand side of
the template that maps an adjective to an adjective. Such labels are
just implementational technicalities that facilitate the retrieval of
the relevant information when a lexical entry is built up from a
template, but they have no role in the matching procedure. For the
present purposes they can entirely be regarded as anonymous variables
that can unify with anything, exactly like their source counterparts.

After transfer, the instances of the templates used in the process are
coindexed in some way, by virtue of their unification with the source
bag items. This is analogous to what happens with bilingual entries in
the translation process.

\subsection{Target bag matching}

The predicate {\tt get\_bag/2} retrieves a bag of lexical items
associated with a derivation. Therefore, {\tt Bag2} and {\tt Bag3}
will contain the bags of lexical items resulting, respectively, from
parsing the target sentence and from transfer.

The crucial step is the matching between the transfer output bag and
the target sentence parse output bag. The predicate {\tt
match\_bags/3} tries to unify the two bags (returning the result in
{\tt Bag4}). A successful unification entails that the parse and
transfer of the source sentence are consistent with the parse of the
target sentence. In other words, the bilingual rules used in transfer
correctly map source lexical items into target lexical
items. Therefore, the lexical equivalences newly established through
this process can be asserted as new bilingual entries.

In the matching process, the order in which the elements are listed in
the figures is irrelevant, since the objects at hand are bags,
i.e. unordered collections. A successful match only requires the
existence of a one-to-one mapping between the two bags, such that:

\begin{enumerate}

\item the respective descriptions, here represented by category
labels, are unifiable;

\item a further one-to-one mapping between the indices in the two bags
is induced.

\end{enumerate}

The following mapping between the transfer output bag (Fig.
\ref{TOB}) and the target sentence parse output bag (Fig.
\ref{TSPOB}) will therefore succeed:

\medskip

\begin{center}

\{{\tt $<$2-1,1$>$,$<$3-2,3$>$,$<$4-3,2$>$,$<$1-4,4$>$,
$<$5-6,5$>$,$<$6-7,7$>$,$<$7-8,6$>$}\}

\end{center}

\medskip

In fact, in addition to correctly unifying the descriptions, it
induces the following one-to-one mapping between the two sets of
indices:

\medskip

\begin{center}

\{{\tt $<$A,0$>$,$<$B,1$>$,$<$I,13$>$}\}

\end{center}

\subsection{Bilingual entries creation}

The rest of the procedure builds up lexical entries for the newly
discovered equivalences and is implementation dependent. First, the
source bag is retrieved in {\tt Bag1}. Then, {\tt make\_be\_info/4}
links together information from the source bag, the target bag
(actually, its unification with the target sentence parse bag) and the
transfer derivation, to construct a list of terms (the variable {\tt
Be}) containing the information to create an entry. Each such term has
the form {\tt be(Sw,Tw,Ti)}, where {\tt Sw} is a list of source words,
{\tt Tw} is a list of target words and {\tt Ti} is a template
identifier. In our example, the following {\tt be/3} terms are
created:

\eenumsentence{

\item{\tt be([fat],[gordo],adj/adj)}

\item{\tt be([man],[hombre],cn/n)}

\item{\tt be([kick,out],[echar],tv+adv/tv)}

\item{\tt be([black],[negro],adj/adj)}

\item{\tt be([dog],[perro],cn/n)}

}

Each {\tt be/3} term is finally turned into a bilingual entry by the
predicate {\tt be\_info\_to\_entries/2}. The following bilingual
entries are created:

\eenumsentence{

\item{\tt 
fat ::@adj(A)\\
$\leftrightarrow$ gordo ::@adj(A).\\
}

\item{\tt 
man ::(D,@count\_noun(C))\\
$\leftrightarrow$ hombre ::(B,@noun(C))\\
$\backslash\backslash$trans\_noun(D,B).\\
}

\item{\tt 
kick ::(I,@trans\_verb(F,G,H))\\
\& out ::@advparticle(F)\\
$\leftrightarrow$\\
echar ::(E,@verb\_acc(F,G,H))\\
$\backslash\backslash$trans\_verb(I,E).\\
}

\item{\tt 
black ::@adj(J)\\
$\leftrightarrow$ negro ::@adj(J).\\
}

\item{\tt 
dog ::(M,@count\_noun(L))\\
$\leftrightarrow$ hombre ::(K,@noun(L))\\
$\backslash\backslash$trans\_noun(M,K).
}

}

If a pre-existing bilingual lexicon is in use, bilingual entries are
prioritized over bilingual templates. Consequently, only new entries
are created, the others being retrieved from the existing bilingual
lexicon. Incidentally, it should be noted that a new entry is an entry
which differs from any existing entry on either side. Therefore,
different entries are created for different senses of the same word,
as long as the different senses have different translations.

\section{Shortcomings and future work}

In matching a pair of bags, two kinds of ambiguity could lead to
multiple results, some of which are incorrect. Firstly, as already
mentioned, a bag could contain two lexical items with unifiable
descriptions (e.g. two adjectives modifying the same noun), possibly
causing an incorrect match. Secondly, as the bilingual template
database grows, the chance of overlaps between templates also
grows. Two different templates or combinations of templates might
cover the same input and output. A case in point is that of a phrasal
verb or an idiom covered by both a single multi-word template and a
compositional combination of simpler templates.

As both potential sources of error can be automatically detected, a
first step in tackling the problem would be to block the automatic
generation of the entries involved when a problematic case occurs, or
to have a user select the correct candidate. In this way the
correctness of the output is guaranteed. The possible cost is a lack
of completeness, when no user intervention is foreseen.

Furthermore, techniques for the automatic resolution of template
overlaps are under investigation. Such techniques assume the presence
of a bilingual lexicon. The information contained therein is used to
assign preferences to competing candidate entries, in two ways.

Firstly, templates are probabilistically ranked, using the existing
bilingual lexicon to estimate probabilities. When the choice is
between single entries, the ranking can be performed by counting the
frequency of each competing template in the lexicon. The entry with
the most frequent template is chosen.

Secondly, heuristics are used to assign preferences, based on the
presence of pre-existing entries related in some way to the candidate
entries. This technique is suited for resolving ambiguities where
multiple entries are involved. For instance, given the equivalence
between {\tt `kick the bucket'} and {\tt `estirar la pata'}, and the
competing candidates

\eenumsentence{

\item \{{\tt kick \& bucket $\leftrightarrow$ estirar \& pata}\}

\item \{{\tt kick $\leftrightarrow$ estirar, bucket $\leftrightarrow$
pata}\}

}

\noindent the presence of an entry `{\tt
bucket~$\leftrightarrow$~balde}' in the bilingual lexicon might be a
clue for preferring the idiomatic interpretation. Conversely, if the
hypothetical entry `{\tt bucket~$\leftrightarrow$~pata}' were already
in the lexicon, the compositional interpretation might be preferred.

Finally, efficiency is also dependant on the restrictiveness of
grammars. The more grammars overgenerate, the more the combinatoric
indeterminacy in the matching process increases. However,
overgeneration is as much a problem for translation as for bilingual
generation. In other words, no additional requirement is placed on the
MT system which is not independently motivated by translation alone.

\section{Conclusion}

The {\em parse-parse-match} approach to automatically building
bilingual lexicons in not novel. Proposals have been put forward,
e.g., by \newcite{Sadler:COLING90} and \newcite{Kaji:COLING92}.

\newcite{Wu:TMI95} points out some possible difficulties of the
parse-parse-match approach. Among them, the facts that ``appropriate,
robust, monolingual grammars may not be available'' and ``the grammars
may be incompatible across languages'' \cite[355]{Wu:TMI95}. More
generally, in bilingual lexicon development there is a tendency to
minimize the need for linguistic resources specifically developed for
the purpose. In this view, several proposals tend to use statistical,
knowledge-free methods, possibly in combination with the use of
existing Machine Readable Dictionaries (see, e.g.,
\newcite{Klavans:MT}, which also contains a survey of related
proposals, pages 195--196).

The present proposal tackles the problem from a different and novel
perspective. The acknowledgment that MT is the main application domain
to which bilingual resources are relevant is taken as a starting
point.  The existence of an MT system, for which the bilingual lexicon
is intended, is explicitly assumed. The potential problems due to the
need for linguistic resources are by-passed by having the necessary
resources available in the MT system. Rather than doing away with
linguistic knowledge, the pre-existing resources of the pursued
application are utilized.

An approach like the present can be most effectively adopted to
develop tools allowing MT systems to automatically build their own
bilingual lexicons. A tool of this sort would use no extra resources
in addition to those already available in the MT system itself. Such a
tool would take a small sample of a bilingual lexicon and use it to
bootstrap the automatic development of a large lexicon. It is worth
noting that the bilingual pairs thus produced would be complete
bilingual entries that could be directly incorporated in the MT
system, with no post-editing or addition of information.

The only requirement placed by the present approach on MT systems is
that they be bi-directional. Therefore, although aimed at the
development of specific applications for specific MT systems, the
approach is general enough to apply to a wide range of MT systems.

\section*{Acknowledgements}

This research was supported by TCC Communications, by a Collaborative
Research and Development Grant from the Natural Sciences and
Engineering Research Council of Canada (NSERC), and by the Institute
for Robotics and Intelligent Systems.  The author would like to thank
Fred Popowich and John Grayson for their comments on earlier versions
of this paper.


\begin{thebibliography}{}

\bibitem[\protect\citename{Buschbeck-Wolf and Dorna}1997]{Buschbeck:RANLP97}
B.~Buschbeck-Wolf and M.~Dorna.
\newblock 1997.
\newblock Using hybrid methods and resources in semantic-based transfer.
\newblock In {\em Proceedings of the International Conference `Recent Advances
  in Natural Language Processing'}, pages 104--111, Tzigov Chark, Bulgaria.

\bibitem[\protect\citename{G\"{u}venir and Tun\c{c}}1996]{Guvenir:CSCSI96}
H.~A. G\"{u}venir and A.~Tun\c{c}.
\newblock 1996.
\newblock Corpus-based learning of generalized parse tree rules for
  translation.
\newblock In G.~McCalla, editor, {\em Advances in Artificial Intelligence ---
  11th Biennial Conference of the Canadian Society for Computational Studies of
  Intelligence}, pages 121--132. Springer, Berlin.

\bibitem[\protect\citename{Kaji \bgroup et al.\egroup }1992]{Kaji:COLING92}
H.~Kaji, Y.~Kida, and Y.~Morimoto.
\newblock 1992.
\newblock Learning translation templates from bilingual text.
\newblock In {\em Proceedings of the 14th International Conference on
  Computational Linguistics}, pages 672--678, Nantes, France.

\bibitem[\protect\citename{Klavans and Tzoukermann}1995]{Klavans:MT}
J.~Klavans and E.~Tzoukermann.
\newblock 1995.
\newblock Combining corpus and machine-readable dictionary data for building
  bilingual lexicons.
\newblock {\em Machine Translation}, 10:185--218.

\bibitem[\protect\citename{Popowich \bgroup et al.\egroup }1997]{Popowich:TMI}
F.~Popowich, D.~Turcato, O.~Laurens, P.~McFetridge, J.~D. Nicholson,
  P.~McGivern, M.~Corzo-Pena, L.~Pidruchney, and S.~MacDonald.
\newblock 1997.
\newblock A lexicalist approach to the translation of colloquial text.
\newblock In {\em Proceedings of the 7th International Conference on
  Theoretical and Methodological Issues in Machine Translation}, pages 76--86,
  Santa Fe, New Mexico, USA.

\bibitem[\protect\citename{Sadler and Vendelmans}1990]{Sadler:COLING90}
V.~Sadler and R.~Vendelmans.
\newblock 1990.
\newblock Pilot implementation of a bilingual knowledge bank.
\newblock In {\em Proceedings of the 13th International Conference on
  Computational Linguistics}, pages 449--451, Helsinki, Finland.

\bibitem[\protect\citename{Turcato \bgroup et al.\egroup
  }1997]{Turcato:RANLP97}
D.~Turcato, O.~Laurens, P.~McFetridge, and F.~Popowich.
\newblock 1997.
\newblock Inflectional information in transfer for lexicalist {M}{T}.
\newblock In {\em Proceedings of the International Conference `Recent Advances
  in Natural Language Processing'}, pages 98--103, Tzigov Chark, Bulgaria.

\bibitem[\protect\citename{Whitelock}1994]{Whitelock:SB}
P.~Whitelock.
\newblock 1994.
\newblock Shake and bake translation.
\newblock In C.J. Rupp, M.A. Rosner, and R.L. Johnson, editors, {\em
  Constraints, Language and Computation}, pages 339--359. Academic Press,
  London.

\bibitem[\protect\citename{Wu}1995]{Wu:TMI95}
D.~Wu.
\newblock 1995.
\newblock Grammarless extraction of phrasal translation examples from parallel
  texts.
\newblock In {\em Proceedings of the Sixth International Conference on
  Theoretical and Methodological Issues in Machine Translation}, pages
  354--372, Leuven, Belgium.

\end{thebibliography}
\end{document}